\documentclass[number]{ReportTemplate}
\usepackage{times}
\usepackage{amssymb}
\usepackage{amsfonts}
\usepackage[fleqn]{amsmath}
\usepackage{mathtools}
\usepackage{amsthm}
\usepackage{bm}
\usepackage{graphicx}
\usepackage{algorithm}
\usepackage{algorithmic}
\usepackage{hyperref}
\usepackage{float,dblfloatfix}

\newcommand{\widebar}{\overline}
\newcommand{\sacii}{SAC$_\text{I\!I}$}
\newcommand{\saci}{SAC$_\text{I}$}

\DeclareMathOperator*{\argmin}{argmin}

\DeclareMathOperator*{\sign}{{sign}}
\newcommand{\pr}{\mathbf{Pr}}

\newcommand{\paa}{PA\!A}

\mathtoolsset{showonlyrefs,showmanualtags}

\newcommand{\expect}[2][]{
    {\mathbb{E}_{#1}[\kern-0.15em[ #2 ]\kern-0.14em]}
    }

\newcommand{\A}{\mathfrak{A}}

\journal{Artificial Intelligence}

\begin{document}
\begin{frontmatter}

\title{The Sampling-and-Learning Framework: \\A Statistical View of Evolutionary Algorithms}
\author{Yang Yu\corref{cor1}}
\ead{yuy@lamda.nju.edu.cn}
\author{Hong Qian}
\ead{qianh@nju.edu.cn}
\cortext[cor1]{Corresponding author}
\address{National Key Laboratory for Novel Software Technology\\
Nanjing University, Nanjing 210023, China}


\begin{abstract}
Evolutionary algorithms (EAs), a large class of general purpose optimization algorithms inspired from the natural phenomena, are widely used in various industrial optimizations and often show excellent performance.
This paper presents an attempt towards revealing their general power from a statistical view of EAs. By summarizing a large range of EAs into the \emph{sampling-and-learning} framework, we show that the framework directly admits a general analysis on the \emph{probable-absolute-approximate} (\paa) query complexity. We particularly focus on the framework with the learning subroutine being restricted as a binary classification, which results in the \emph{sampling-and-classification} (SAC) algorithms. With the help of the learning theory, we obtain a general upper bound on the \paa{} query complexity of SAC algorithms. We further compare SAC algorithms with the uniform search in different situations. Under the \emph{error-target independence} condition, we show that SAC algorithms can achieve polynomial speedup to the uniform search, but not super-polynomial speedup. Under the \emph{one-side-error} condition, we show that super-polynomial speedup can be achieved. This work only touches the surface of the framework. Its power under other conditions is still open.
\end{abstract}

\end{frontmatter}


\section{Introduction}

In many practical optimization problems, the objective functions are hidden or too complicated to be analyzed. Under this kind of circumstances, direct optimization algorithms are appealing, which follows the trial-and-error style with some heuristics. Evolutionary algorithms (EAs) \cite{back:96} are a large family of such algorithms. The family includes genetic algorithms \cite{goldberg1989genetic}, evolutionary programming \cite{koza1994genetic}, evolutionary strategies \cite{beyer.schwefel.02}, and also covers other nature-inspired heuristics including particle swarm optimization \cite{kennedy1995particle}, ant colony optimization \cite{dorigo.etal.96}, estimation of distribution algorithms \cite{larranaga2002estimation}, etc. 

Theoretical studies of EAs have been developed rapidly in the recent decades, particularly noticeable of the blooming of running time analysis \cite{neumann.witt.10,auger.doerr.11,jansen13}. With the development of several analysis techniques (e.g. \cite{he2001drift,yu2008new,doerr2011multiplicative,sudholt2011general}),  EAs have been theoretically investigated on problems from simple synthetic ones (e.g. \cite{droste2002analysis}) to combinatorial problems (e.g. \cite{Scharnow.etal.PPSN02}) as well as NP-hard problems (e.g. \cite{yu.etal.aij12}). During these analyses, effects of EAs components have been disclosed \cite{yao.12}, including the crossover operators (e.g. \cite{jansen2002analysis,lehre2008crossover,DoerrJKNT13,qian.etal.aij13}), the population size (e.g. \cite{jansen2005choice,storch2008choice,witt2008population,chen2009new}), etc. Measures of the performance also have developed to cover the approximation complexity (e.g. \cite{he_yao_03_cec,Friedrich.etal.GECCO07,
yu.etal.aij12,Lai.etal.tec14}), the fixed-parameter complexity (e.g. \cite{KratschN.Algo13,SuttonN.aaai12}), the complexity under fixed-budget computation \cite{Jansen.gecco12}, etc. While most of these analyses studied instances of EAs on problem cases, general performance analysis may even be more desired, as the application of EAs is nearly unlimited. The famous No-Free-Lunch Theorem \cite{wolpert1997no} used a quite general framework of EAs and gave a general conclusion that any two EAs are with the same performance (at least on discrete domains) given no prior knowledge of the problem distribution, of which the general running time is exponential \cite{yu2008new}. When the complexity of a problem class is bounded, a general convergence lower bound can be derived for a class of EAs \cite{fournier2011lower}. For more general EAs, the Black-Box model can derive the best possible performance \cite{Droste.etal.foga02,Anil.Wiegand.foga09,Lehre.witt.algorithmica12,doerr.winzen.11}. We have learned that a general performance analysis relies on a general framework of EAs.

It has been noticed that various implementations of EAs share a common structure that consists of a cycle of sampling and model building \cite{zlochin.etal.aor04}. In this work, we propose to study the \emph{sampling-and-learning} (SAL) framework. EAs commonly employ some heuristic to reproduce solutions, which is captured by the sampling step of SAL; and they also distinguish the quality of the reproduced solutions to guide the next sampling (e.g., genetic algorithms remove a portion of the worst solutions), which is captured by the learning step of SAL. The SAL framework can simulate a wide range of EAs as well as other heuristic search methods, by specifying the sampling and the learning strategies.

We evaluate this framework by the probable-absolute-approximate (\paa{}) query complexity. \paa{} complexity counts the number of fitness evaluations before reaching to an approximate solution with a probability, which is close to the intuitive evaluation of EAs in practice.
We show that the SAL framework immediately admits a general \paa{} upper bound. For a specific version of SAL that uses classification algorithms, named the SAC algorithms, we obtain a tighter \paa{} upper bound by incorporating the learning theory results. Further comparing with the uniformly random search, we disclose that, under the \emph{error-target independence} condition, SAC algorithms can polynomially reduce the complexity of the uniform search, but not super-polynomially; while the  \emph{one-side-error} condition further allows a super-polynomial improvement. This study shows that the classification error is an important effecting factor, which was not noticed before. We also notice that a good learning algorithm may not be necessary for a good SAL algorithm.

The rest of this paper is organized as follows: Section II introduces the SAL framework. In Section III, we compare the SAC algorithms, a specific version of the SAL framework, with the uniform search. Finally, Section IV concludes the paper.

\section{The Sampling-and-Learning Framework}

In this paper, we consider general minimization problems $f$. We always denote $X$ as the whole solution space which an algorithm will search among. In the analysis of this paper, we consider $X \subseteq \mathbb{R}^n$ is a compact set (in the Euclidean space, the compact set is equivalent to the bounded and closed set) and $f: X\to \mathbb{R}$ is a continuous function. Thus there must exist at least one solution $x^{\ast} \in X$ such that $f(x^{\ast}) = \min_{x \in X}f(x)$. We use $D$ to denote sub-regions of $X$ and define $|D|=\int_{D}{1}\mathrm{d}x$. For the sake of convenience for the analysis, we assume without loss of generality that $|X| = 1$ since $X$ is a bounded and closed set. Denote $D_{\alpha} = \{x\in X|f(x)\leq {\alpha}\}$ for any scaler ${\alpha}$, $\mathcal{U}_X$ as the uniform distribution over $X$, $\mathcal{T}$ and $\mathcal{D}$ as the probability distributions. Besides, by $\texttt{poly}(\cdots)$, we mean the set of all polynomials with the related variables, and by $\texttt{superpoly}(\cdots)$, we mean the set of all functions that grow faster than any function in $\texttt{poly}(\cdots)$ with the related variables.
\begin{definition}[Minimization Problem]
A minimization problem consists of a continuous solution space $X$ and a continuous function $f: X\to \mathbb{R}$, where $X \subseteq \mathbb{R}^n$ and $X$ is a compact set. The goal is to find a solution $x^*\in X$ such that $f(x^*)\leq f(x)$ for all $x\in X$.
\end{definition}

Since $X$ is a compact set and $f$ is a continuous function, there must exist one solution $x' \in X$ such that $f(x') = \max_{x \in X}f(x)$. Namely, $f$ is bounded in $[f(x^*),f(x')]$. Therefore, in the rest of the paper, we assume without loss of generality that the value of $f$ is bounded in $[0,1]$, i.e., $\forall x\in X:f(x)\in [0,1]$. Given an arbitrary function $g$ with bounded value range over the input domain, the bound can be implemented by a simple normalization $f(x)=\frac{g(x)-g(x^*)}{\max_{x'} g(x')-g(x^*)}$. Thus we assume in the rest of this paper that every minimization problem has its minimum value $0$.

In real-world applications, we expect EAs to achieve some good enough solutions with a not quite small probability, which corresponds to approximation (e.g. \cite{yu.etal.aij12}) and probabilistic performance (e.g. \cite{Zhou.etal.TCS12}). Combining the two, we study the \emph{probable-absolute-approximate} (\paa{}) query complexity, which is the number of fitness evaluations that an algorithm takes before reaching an approximate quality, as defined in Definition \ref{paa}. The \paa{} query complexity closely reflects our intuitive evaluation of EAs in practice.

\begin{definition}[Probable-Absolute-Approximate Query Complexity]\label{paa}
Given a minimization problem $f$, an algorithm $\mathcal A$, and any $0<\delta<1$ as well as any approximation level $\alpha^* > 0$, then the probable-absolute-approximate (\paa{}) query complexity is the number of calls to $f(\cdot)$ such that, with probability at least $1-\delta$, $\mathcal A$ finds a solution $x$ with $f(x)\leq \alpha^*$.
\end{definition}

\subsection{The General Framework}

Most EAs share a common trial-and-error structure with several important properties:\\[-1.8em]
\begin{itemize}
\item[a)] directly access the solution space, generate solutions, and evaluate the solutions;\\[-1.8em]
\item[b)] the generation of new solutions depends only on a short history of past solutions;\\[-1.8em]
\item[c)] both ``global'' and ``local'' heuristic operators are employed to generate new solutions.\\[-1.8em]
\end{itemize}
We present a sampling-and-learning (SAL) framework in Algorithm \ref{framework} to capture these properties. The SAL framework starts from a random sampling in Step 1 like all EAs. Steps 2 and 13 record the best-so-far solutions throughout the search. SAL follows a cycle of learning and sampling stages. In Step 7, it learns a hypothesis $h_t$ (i.e., a mapping from $ X$ to $\mathbb{R}$) via the learning algorithm $\mathcal{L}$. Note that the learning algorithm allows to take the current data set $T_t$, the last data set $T_{t-1}$, and the last hypothesis $h_{t-1}$ into account. Different EAs may make different use of them. Step 8 initializes the sample set for the next iteration. The sample set can be initialized as an empty set, or to preserve some good solutions from the previous iteration. In Steps 9 to 12, it samples from the distribution transformed from the hypothesis as well as from the whole solution space balanced by a probability. The distribution $\mathcal{T}_{h_t}$ implies the potential good regions learned by $h_t$.

\begin{algorithm}[t!]
\caption{The sampling-and-learning (SAL) framework}
\label{framework}
\begin{algorithmic}[1]
\REQUIRE ~~\\
{${\alpha^*}>0$:} Approximation level\\
{$T\in \mathbb{N}^+$:} Number of iterations \\
{$m_0,\ldots, m_T \in \mathbb{N}^+$:} Number of samples\\
{$\lambda \in [0,1]$:} Balancing parameters\\
{$\mathcal{L}$:} Learning algorithm\\
{$\mathcal{T}$:} Distribution transformation of hypothesis
\ENSURE ~~\\
\STATE Collect $S_0=\{x_1,\ldots, x_{m_0}\}$ by i.i.d. sampling from the uniform distribution over $ X$
\STATE $\tilde{x} = \argmin_{x\in S_0} f(x)$
\STATE Initialize the hypothesis $h_0$
\STATE $T_0=\emptyset$
\FOR{$t=1$ to $T$}
    \STATE Construct $T_t=\{(x_1,y_1),\ldots,(x_{m_{t-1}}, y_{m_{t-1}})\}$,\\\qquad\qquad\qquad\quad\quad where $x_i\in S_{t-1}$ and $y_i=f(x_i)$
    \STATE $h_t = \mathcal{L}(T_t, T_{t-1}, h_{t-1},t)$, the learning step
	\STATE Initialize $S_t$ from $T_t$
	\FOR{$i=1$ to $m_t$}
		\STATE Sample $x_i$ from
		$\begin{cases}
			\mathcal{T}_{h_t},& \text{with probability } \lambda\\
			\mathcal{U}_{ X},& \text{with probability } 1-\lambda
		\end{cases}
		$
		\STATE $S_t=S_t\cup \{x_i\}$
	\ENDFOR
	\STATE $\tilde{x} = \argmin_{x\in S_t\cup \{\tilde{x}\}} f(x)$
\ENDFOR
\RETURN  $\tilde{x}$
\end{algorithmic}
\end{algorithm}
\begin{figure}[t!]\vspace{-3em}
\end{figure}

It should be noted that the SAL framework is not a concrete optimization algorithm but an abstract summary of a range of EAs, nor does the learning stage of the framework imply an accurate learning. We explain in the following how we could mimic several different EAs by the SAL framework. It is noticeable that the explanation is not a rigorous proof, but an intuitive illustration that the SAL framework can correspond to various implementations. 

The genetic algorithms (GAs) \cite{goldberg1989genetic} deal with discrete solution spaces consisting of solutions represented as a vector of  vocabulary. The element-wise mutation operator changes every element of a solution to a randomly selected word from the vocabulary with a probability. Converting this operation probability to the probability of generating a certain solution, let $P_m(x'|x)$ be the probability of generating the solution $x'$ from $x$ via the element-wise mutation, thus $P_m(x'|x)=(\frac{p}{|V|-1})^{\|x'-x\|_H}(1-p)^{n-\|x'-x\|_H}$, where $n$ is the length of the solution, $|V|$ is the vocabulary size, $\|\cdot \|_H$ is the Hamming distance, and $p$ is the probability of changing the element that is commonly $\frac{1}{n}$. It is easy to calculate that $P_m(x'|x)$ is $\frac{1}{\texttt{poly(n)}}$ only when $\|x'-x\|_H$ is a constant (and otherwise $P_m(x'|x)=\frac{1}{\texttt{superpoly(n)}}$). Given any set of solutions $S=\{x_1, x_2, \ldots, x_m\}$, we divide the search space into two sets that $X_{poly}(S)=\{x\in X\mid \exists x'\in S: \|x-x'\|_H =O(1)\}$ and $X_{super}(S)=X-S_{poly}(S)$. SAL can simulate the GA as that, for every population $S$ of the GA, SAL learns the hypothesis $h$ that circles the area $X_{poly}(S)$, and uses $\mathcal T_{h}$ as $\mathcal T_{h}(x) = \frac{\sum_{x\in S} P_m(x'|x)}{\sum_{x''\in X_{poly}(S)}\sum_{x\in S} P_m(x''|x)}$ for solutions in $X_{poly}(S)$. And for the area $X_{super}(S)$, SAL uses the uniform distribution to approximate the sampling with super-polynomially small probability. In this way, SAL can mimic the behavior of the GA. We have discussed a simplified GA. Most GAs also employ the crossover operators, which is a kind of local search operator and thus the resulting distribution can be compiled into the local distribution. Many GAs also employ a probabilistic selection, which can be simulated by selecting the initial solution set $S_t$ in the same way.

It has been argued that model-based search algorithms including the estimation of distribution algorithms (EDAs) \cite{larranaga2002estimation}, the ant colony optimization algorithms (ACOs) \cite{dorigo.etal.96}, the cross-entropy method \cite{crossentropy.04} can be unified in the sampling and model building framework \cite{zlochin.etal.aor04}, which respectively correspond to the sampling and learning steps in the SAL framework. The particle swarm optimization algorithms (PSOs) \cite{kennedy1995particle} is particularly interesting since the simulation is perhaps the most sophisticated. A PSO algorithm maintains a set of ``flying'' particles each with a location (representing a solution) and a velocity vector. The location of a particle in the next iteration is determined by its current location and current velocity, and the velocity is updated by the current velocity and the locations of the ``globally'' and ``personally'' best particles. To simulate a PSO, a SAL algorithm needs to use the initial hypothesis resulting the same sampling distribution as that from the initial velocity. Let $S_t$ be an ordered set to contain the globally best particle and the personally best particles in Step 8. The learning algorithm in the SAL algorithm can be set to utilize the current data set and the last data set to recover the velocity, and utilize the last hypothesis and the globally and personally best particles recorded through $S_t$ to generate the new hypothesis that simulates the movement of particles in the PSO.

Overall, the SAL framework captures the trial-and-error structure as well as the global--local search balance, while leaving the details of the local sampling distribution being implemented by different heuristics.

The SAL framework directly admits a general upper bound of the \paa{} query complexity, as stated in Theorem \ref{lem-sal}.

\begin{theorem}\label{lem-sal}
For any minimization problem $f$ and any approximation level $\alpha^*>0$, with probability at least $1-\delta$, a SAL algorithm will output a solution $x$ with $f(x)\leq {\alpha^*}$ using $m_{\Sigma}$ number of queried samples bounded from above by
\begin{align}
\!\!O\!\left(\!m_0\!+\!\max\Big\{\!\frac{1}{(1-\lambda)\pr_u + \lambda \widebar{\pr}_h} \ln \frac{1}{\delta},\sum\nolimits_{t=1}^T\! m_{\pr_{h_t}} \Big\}\!\right),\!\!
\end{align}
where $\pr_u = \int_{D_{{\alpha^*}}} \!\mathcal{U}_{X}(x)\, \mathrm{d} x  $ is the success probability of uniform sampling,
\begin{align}
\widebar{\pr}_h = \frac{\sum_{t=1}^{T} m_t \cdot \pr_{h_t}}{\sum_{t=1}^{T} m_t} = \frac{\sum_{t=1}^{T} m_t \cdot \int_{D_{{\alpha^*}}} \!\mathcal T_{h_t}(x)\, \mathrm{d}x}{\sum_{t=1}^{T} m_t} 
\end{align} 
is the average success probability of sampling from the learnt hypothesis, $m_{\pr_{h_t}}$ is the required sample size realizing $\pr_{h_t}$, and $D_{{\alpha^*}} = \{ x \in X|f(x) \leq \alpha^* \}$.
\end{theorem}
\begin{myproof}
$m_0$ is the initial sample size. In every iteration, we need $m_{\pr_{h_t}}$ samples to realize the probability $\pr_{h_t}$ (generally the higher the probability the larger the sample size, but it depends on the concrete implement of the algorithm), thus $\sum_{t=1}^T m_{\pr_{h_t}}$ number of samples is naturally required. We prove the rest of the bound.

Let's consider the probability that after $T$ iterations, the SAL algorithm outputs a bad solution $x$ such that $f(x)>{\alpha^*}$. Since the $x$ is the best solution among all sampled examples, the probability is the intersection of events that every step of the sampling does not generate such a good solution. \\
1. For the sampling from uniform distribution over the whole solution space $ X$, the probability of failure is $1-\pr_u$. \\
2. For the sampling from the learnt hypothesis $h_t$ according to the distribution $\mathcal T_{h_t}$, the probability of failure is denoted as $1-\pr_{h_t}$.\\
Since every sampling is independent, we can expand the probability of overall failures, i.e., for any solution $x$ belongs to the all sampled examples,
\begin{align}
& \pr(f(x)> {\alpha^*})  \\
& = (1-\pr_u)^{m_0} \cdot\\
&\quad \prod_{t=1}^T  \sum_{i=0}^{m_t}\! {m_t\! \choose\! i}(1-\lambda)^i\lambda^{m_t-i} (1\!-\!\pr_u)^i (1\!-\! \pr_{h_t})^{m_t-i} \\
& = (1-\pr_u)^{m_0} \prod\nolimits_{t=1}^T  \left( 1-(1-\lambda)\pr_u-\lambda \pr_{h_t}\right)^{m_t}\\
& \leq e^{- \pr_u\cdot m_0}  \prod\nolimits_{t=1}^T e^{-\left( (1-\lambda) \pr_u m_t + \lambda \pr_{h_t} m_t \right)}\\
& = e^{- \left(\pr_u\cdot m_0 + (1-\lambda) \sum_{t=1}^T \pr_u m_t + \lambda \sum_{t=1}^T \pr_{h_t} m_t \right)}\\
& \leq e^{- \left((1-\lambda) \sum_{t=1}^T \pr_u m_t + \lambda \sum_{t=1}^T \pr_{h_t} m_t \right)}\\
& = e^{ - \left( (1-\lambda)\pr_u + \lambda \widebar{\pr}_h \right) \sum_{t=1}^{T}{m_t} },
\end{align}
where the first inequality is by $(1-x)\leq e^{-x}$ for $x\in [0,1]$.

In order that $\pr(f(x)> {\alpha^*})<\delta$, we let $e^{ - \left( (1-\lambda)\pr_u + \lambda \widebar{\pr}_h \right) \sum_{t=1}^{T}{m_t} } < \delta$, which solves that $\sum_{t=1}^{T}{m_t}=O\Big(\frac{1}{(1-\lambda)\pr_u+\lambda\widebar\pr_h}\ln\frac{1}{\delta}\Big)$.
\end{myproof}

\subsection{The Sampling-and-Classification Algorithms}

To further unfold the unknown term $\widebar{\pr}_h$ in Theorem \ref{lem-sal}, we focus on a simplified version of the SAL framework that employs a classification algorithm in the learning stage. We call this type of algorithms as the \emph{sampling-and-classification} (SAC) algorithms. In the learning stage of a SAC algorithm, as described in Algorithm \ref{sac}, the learning algorithm first uses a threshold to transform the data set into a binary labeled data set, and then invokes the classification algorithm to learn from the binary data set. $\sign[\cdot]$ is defined as $\sign[v]=+1 \text{ if } v \geq 0 \text{ and } -1 \text{ if } v<0$. Note that SAC algorithms use the current data set $T$ in the learning algorithm, but not the last data set $T'$ and the last hypothesis $h'$. Putting Algorithm \ref{sac} into the framework of Algorithm \ref{framework}, we always set $S_t=\emptyset$ for SAC, and $\mathcal T_h$ will be some distribution over the positive area of $h$.

\begin{algorithm}[h!]
\caption{Learning sub-procedure for the sampling-and-classification (SAC) algorithms}
\label{sac}
\begin{algorithmic}[1]
\REQUIRE ~~\\
{$T,T',h',t$:} The input variables\\
{${\alpha}_1>\ldots>\alpha_t$:} Preset threshold parameters\\
{$\mathcal{C}$:} Classification algorithm
\ENSURE ~~\\
    \STATE Construct $B=\{(x_1,z_1),\ldots,(x_{|T|}, z_{|T|})\}$ from that,\\\quad for all $i$ and all $(x_i, y_i) \in T$, $z_i=\sign[\alpha_t-y_i]$ 
    \STATE $h = \mathcal{C}(B)$
	\RETURN $h$
\end{algorithmic}
\end{algorithm}

By these specifications, we can have a general \paa{} performance for SAC algorithms. According to Theorem \ref{lem-sal}, we need to estimate a lower bound of $\widebar\pr_h$, i.e., how likely the distribution $\mathcal T_{h_t}$ will lead to a good solution. Recall $D_{\alpha} = \{x\in X|f(x)\leq {\alpha}\}$ for any scaler $0<{\alpha}<1$. Denote $D_h = \{x\in X|h(x) = +1\}$ for any hypothesis $h$,  $\mathcal U_{D_h}$ as the uniform distribution over $D_h$, and $D_{KL}$ as the Kullback-Leibler (KL) divergence. KL-divergence measures how difference one distribution departs from another one. For probability distributions $P$ and $Q$ of two continuous random variables, $D_{KL}(P||Q) = \int_{-\infty}^{+\infty}{\ln\left(\frac{p(x)}{q(x)}\right)p(x)\mathrm{d}x}$, where $p(x)$ and $q(x)$ are the probability densities of $P$ and $Q$. Let $\Delta$ denote the symmetric difference operator of two sets. We have a lower bound of the success probability as in Lemma \ref{lemma-localsuccess}.

\begin{lemma}\label{lemma-localsuccess}
For any minimization problem $f$, any approximation level ${\alpha^*}>0$, any hypothesis $h$, the probability that a solution sampled from an arbitrary distribution $\mathcal T_h$ defined on $D_h$ will lead to a solution in $D_{\alpha^*}$ is lower bounded as
\begin{align}
\pr_h \geq \frac{|D_{\alpha^*}\cap D_h|}{|D_h|} - |D_{\alpha^*}\cap D_h|\sqrt{\frac{1}{2}D_{KL}(\mathcal T_h\|\mathcal U_{D_h})}
\end{align}
\end{lemma}
\begin{myproof}
	Let $I[\cdot]$ denote the indicator function, namely, $I[\texttt{true}]=1$ and $I[\texttt{false}]=0$. The proof starts from the definition of the probability,
	\begin{align}
		&\pr_h = \int_{D_h} \mathcal T_h (x)\cdot I[x\in D_{\alpha^*}] \mathrm d x\\
		&= \int_{D_h} (\mathcal T_h(x)-\mathcal U_{D_h}(x)+\mathcal U_{D_h}(x))\cdot I[x\in D_{\alpha^*}] \mathrm d x\!\!\\
		& = \frac{|D_{\alpha^*}\cap D_h|}{|D_h|} + \int_{D_h} (\mathcal T_h(x)-\mathcal U_{D_h}(x))\cdot I[x\in D_{\alpha^*}] \mathrm d x\!\!\\
		& \geq \frac{|D_{\alpha^*}\cap D_h|}{|D_h|} - \!\!\int_{D_h} \!\!\sup\limits_{x'}|\mathcal T_h(x')\!-\!\mathcal U_{D_h}(x')|\!\cdot\! I[x\!\in\! D_{\alpha^*}] \mathrm d x\!\!\!\!\!\!\\
		& \geq \frac{|D_{\alpha^*}\cap D_h|}{|D_h|} - \sqrt{\frac{1}{2}D_{KL}(\mathcal T_h\|\mathcal U_{D_h})}\int_{D_h}  I[x\in D_{\alpha^*}] \mathrm d x\!\!\\
		& = \frac{|D_{\alpha^*}\cap D_h|}{|D_h|} - |D_{\alpha^*}\cap D_h|\sqrt{\frac{1}{2}D_{KL}(\mathcal T_h\|\mathcal U_{D_h})},
	\end{align}
	where the last inequality is by Pinsker's inequality.
\end{myproof}

We cannot pre-determine $D_h$, but we know that $h$ is derived by a binary classification algorithm from a data set which is labeled according to the threshold parameter $\alpha$. For the binary classification, we know that the generalization error, which is the expected misclassification rate, can be bounded above by the training error, which is the misclassification rate in the seen examples, as well as the generalization gap involving the complexity of the hypothesis space \cite{Kearns.94}, as in Lemma \ref{lemma-classification}. The $VC(\mathcal H)$ is the VC-dimension measuring the complexity of $\mathcal H$.

\begin{lemma}[\cite{Kearns.94}]\label{lemma-classification}
Let $\mathcal{H}=\{h: X \to \{-1,+1\}\}$ be the hypothesis space containing a family of binary classification functions and $VC(\mathcal{H})=d$, if there exist $m$ samples i.i.d. from $ X$ according to some fixed unknown distribution $\mathcal D$, then, $\forall\ h \in \mathcal{H}$ and $\forall\ 0<\eta<1$, the following upper bound holds true with probability at least $1-\eta$:
\begin{align}
\epsilon_{\mathcal D} \leq \hat{\epsilon}_{\mathcal D}+
\sqrt{8m^{-1}\big(d\log{(2emd^{-1})}+\log{(4\eta^{-1})}\big)}
\end{align}
where $\epsilon_{\mathcal D}$ is the expected error rate of $h$ over $\mathcal D$ and $\hat\epsilon_{\mathcal D}$ is the error rate in the sampled examples from ${\mathcal D}$, and when $\hat{\epsilon}_{\mathcal D}=0$,
\begin{align}
\epsilon_{\mathcal D} \leq 2m^{-1}\big(d\log{(2emd^{-1})}+\log{(2\eta^{-1})}\big).
\end{align}
\end{lemma}

Again by Pinsker's inequality, we know that the error $\epsilon_\mathcal D$ under the distribution $\mathcal D$ can be converted to the error $\epsilon_{\mathcal U}$ under the uniform distribution, as
\begin{align}
\epsilon_{\mathcal U} & \leq \frac{\epsilon_{\mathcal D}}{1 - \sqrt{\frac{1}{2}D_{KL}(\mathcal D\|\mathcal U)} }\\
& \leq \frac{\hat{\epsilon}_{\mathcal D}+\sqrt{8m^{-1}\big(d\log{(2emd^{-1})}+\log{(4\eta^{-1})}\big)}} {1 - \sqrt{\frac{1}{2}D_{KL}(\mathcal D\|\mathcal U)} },
\end{align}
where we only take the event that the generalization inequality holds with probability $1-\eta$ into account.
For simplicity, we denote the right-hand part as $\Psi_{\hat{\epsilon}_{\mathcal D}, d, D_{KL}(\mathcal D\|\mathcal U)}^{m, \eta}$, which decreases with $m$ and $\eta$, and increases with $\hat{\epsilon}_{\mathcal D}$, $d$, and $D_{KL}(\mathcal D\|\mathcal U)$.

We can use this inequality to eliminate the $D_h$ in Lemma \ref{lemma-localsuccess}. In every iteration of SAC algorithms, there are $m_t$ samples collected, which make the error of $h_t$ bounded.

\begin{theorem}\label{theorem-sac}
For any minimization problem $f$, any constant $0<\eta<1$, and any approximation level ${\alpha^*}>0$, the average success probability of sampling from the learnt hypothesis of any SAC algorithm is lower bounded as
\begin{align}
\widebar\pr_{h} \geq \ & \frac{1-\eta}{\sum_{t=1}^T m_t} \sum_{t=1}^T m_t \left(\frac{|D_{\alpha^*}|-2\Psi_{\hat{\epsilon}_{\mathcal D_t}, d, D_{KL}(\mathcal D_t\|\mathcal U_{ X})}^{m_t, \eta}}{|D_{\alpha_t}|+\Psi_{\hat{\epsilon}_{\mathcal D_t}, d, D_{KL}(\mathcal D_t\|\mathcal U_{ X})}^{m_t, \eta}}
 - |D_{\alpha^*}|\sqrt{\frac{1}{2}D_{KL}(\mathcal T_{h_t}\|\mathcal U_{D_{h_t}})}\right),
\end{align}
where $\mathcal{D}_t = \lambda \mathcal T_{h_{t}} + (1-\lambda) \mathcal U_{ X}$ is the sampling distribution at iteration $t$, $\hat\epsilon_{\mathcal D_t}$ is the training error rate of $h_t$, $d$ is the VC-dimension of the learning algorithm.
\end{theorem}
\begin{myproof}
By set operators,
\begin{align}
	& |D_{{\alpha^*}} \cap D_{h_t}| = |D_{{\alpha^*}} \cup D_{h_t}| - |D_{{\alpha^*}} \Delta D_{h_t}|\\
	& \geq  |D_{{\alpha^*}} \cup D_{h_t}| - |D_{{\alpha^*}} \Delta D_{\alpha_t}| - |D_{\alpha_t} \Delta D_{h_t}|\\
	& =  |D_{{\alpha^*}} \cup D_{h_t}| - |D_{{\alpha^*}} \Delta D_{\alpha_t}| - \epsilon_{\mathcal U_X,t}\\
	& = |D_{{\alpha^*}} \cup D_{h_t}| + |D_{{\alpha^*}}| - |D_{\alpha_t}| - \epsilon_{\mathcal U_X,t},
\end{align}
where $\Delta$ is the symmetric difference operator of two sets and $\epsilon_{\mathcal U_X,t}$ is the expected error rate of $h_t$ under $\mathcal U_X$. The first inequality is by the triangle inequality, and the last equation is by that $D_{\alpha^*}$ is contained in $D_{\alpha_t}$.

Since
$
\big| |D_{h_t}| - |D_{\alpha_t}|\big| \leq |D_{h_t} \Delta D_{\alpha_t}| = \epsilon_{\mathcal U_X,t},
$
we can bound $|D_{h_t}|$ as
$
|D_{\alpha_t}| +\epsilon_{\mathcal U_X,t} \geq |D_{h_t}| \geq |D_{\alpha_t}| -\epsilon_{\mathcal U_X,t}.$

 Now, we can apply Lemma \ref{lemma-localsuccess}, and the success probability of sampling from $D_{h_t}$ is lower bounded as
\begin{align}
	&\pr_{h_t} \geq \frac{|D_{\alpha^*}\cap D_{h_t}|}{|D_{h_t}|} - |D_{\alpha^*}\cap D_{h_t}|\sqrt{\frac{1}{2}D_{KL}(\mathcal {T}_{h_t}\|\mathcal U_{D_{h_t}})}\\
    & \geq \frac{1}{|D_{h_t}|} \cdot (|D_{{\alpha^*}} \cup D_{h_t}| + |D_{{\alpha^*}}| - |D_{\alpha_t}| - \epsilon_{\mathcal U_X,t})  - |D_{{\alpha^*}}|\cdot \sqrt{\frac{1}{2}D_{KL}(\mathcal T_{h_t}\|\mathcal U_{D_{h_t}})}\\
    & \geq  \frac{1}{|D_{h_t}|} \cdot (|D_{h_t}| + |D_{{\alpha^*}}| - |D_{\alpha_t}| - \epsilon_{\mathcal U_X,t})  - |D_{{\alpha^*}}| \cdot \sqrt{\frac{1}{2}D_{KL}(\mathcal T_{h_t}\|\mathcal U_{D_{h_t}})}\\
	& \geq \frac{|D_{\alpha^*}|-2\epsilon_{\mathcal U_X,t}}{|D_{\alpha_t}|+\epsilon_{\mathcal U_X,t}} - |D_{\alpha^*}|\sqrt{\frac{1}{2}D_{KL}(\mathcal T_{h_t}\|\mathcal U_{D_{h_t}})}.
\end{align}
Substituting this lower bound and the probability $1-\eta$ of the generalization bound into $\widebar\pr_h$ obtains the theorem.
\end{myproof}

Combining Theorem \ref{lem-sal} and Theorem \ref{theorem-sac} results an upper bound on the sampling complexity of SAC algorithms. Although the expression looks sophisticated, it can still reveal relative variables that generally effect the complexity. One could design various distributions for $\mathcal T_h$ to sample potential solutions, however, without any a priori knowledge, the uniform sampling is the best in terms of the worst case performance. Meanwhile, without any a priori knowledge, a small training error at each stage from a learning algorithm with a small VC-dimension can also improve the performance.

\section{SAC Algorithms v.s. Uniform Search}

When EAs are applied, we usually expect that they can achieve a better performance than some baselines. The uniform search can serve as a baseline, which searches the solution space always by sampling solutions uniformly at random. In other words, the uniform search is the SAL algorithm with $\lambda=0$. In this section, we study the performance of SAC algorithms relative to the uniform search.

SAC algorithms will degenerate to uniform search if $\lambda=0$. Thus, it is easy to know that the \paa{} query complexity of uniform search is
\begin{align}
\Theta\left(\frac{1}{\pr_u } \cdot \ln \frac{1}{\delta}\right).
\end{align}
Contrasting this with Theorem 1, we can find that how much a SAC algorithm improves from the uniform search depends on the average success probability $\widebar{\pr}_h$ that relies on the learnt hypothesis. A SAC algorithm is not always better than the uniform search. Without any restriction, $\widebar{\pr}_h$ can be zero and thus the SAC algorithm is worse. 
We are then interested in investigating the conditions under which SAC algorithms can accelerate from the uniform search.

\subsection{A Polynomial Acceleration Condition}

\begin{cond}[Error-Target Independence] In SAC algorithms, for any $t$ and any approximation level ${\alpha^*}>0$, when sampling a solution $x$ from $\mathcal U_{ X}$, the event $x\in D_{h_t}\Delta D_{\alpha_t}$ and the event $x\in D_{\alpha^*}$ are independent.
\end{cond}

We call SAC algorithms that are under the error-target independence condition as \saci{} algorithms. The condition is defined using the independence of random variables. From the set perspective, it is equivalent with
\begin{align}
|D_{\alpha^*} \cap (D_{\alpha_{t}} \Delta D_{h_{t}})|=|D_{\alpha^*}|\cdot |(D_{\alpha_{t}} \Delta D_{h_{t}})|.
\end{align}
Under the condition, we can bound from below the probability of sampling a good solution, as stated in Lemma \ref{lem-saci}.

\begin{lemma}\label{lem-saci}
For \saci{} algorithms, it holds for all $t$ that
\begin{align}
\frac{|D_{\alpha^*}\cap D_{h_t}|}{|D_{h_t}|} \geq \frac{|D_{\alpha^*}|(1-\epsilon_{\mathcal U_X,t})}{|D_{\alpha_t}|+\epsilon_{\mathcal U_X,t}},
\end{align}
where $\epsilon_{\mathcal U_X,t}$ is the expected error rate of $h_t$ under $\mathcal U_X$.
\end{lemma}
\begin{myproof}
For the numerator,
\begin{align}
& |D_{\alpha^*}\cap D_{h_t}| = |D_{\alpha^*}| - |D_{\alpha^*} \cap (D_{\alpha_{t}} \Delta D_{h_{t}})|\\
& = |D_{\alpha^*}| - |D_{\alpha^*}|\cdot|D_{\alpha_{t}} \Delta D_{h_t}| \\
& \geq |D_{\alpha^*}|(1-\epsilon_{\mathcal U_X,t}),
\end{align}
where the first equation is by $D_{\alpha^*} \subseteq D_{\alpha_{t}}$, and the second equality is by the error-target independence condition.

For the denominator, we consider the worst case that all errors are out of $D_{h_t}$ and thus
$|D_{h_t}| \leq |D_{\alpha_t}| + \epsilon_{\mathcal U_X,t}$.
\end{myproof}

Similar to Theorem \ref{theorem-sac}, we can bound from below the average success probability of sampling from the positive area of the learnt hypothesis,
\begin{align}
\widebar{\pr}_h \geq &\ \frac{1-\eta}{\sum_{t=1}^{T} m_t} \sum\nolimits_{t=1}^{T} m_t \Big(\frac{|D_{\alpha^*}|(1-\epsilon_{\mathcal U_X,t})}{|D_{\alpha_t}|+\epsilon_{\mathcal U_X,t}}  - |D_{\alpha^*}|\sqrt{\frac{1}{2}D_{KL}(\mathcal T_{h_t}\|\mathcal U_{D_{h_t}})}\Big).
\end{align}

We compare the uniform search with the \saci{} algorithms using uniform sampling within $D_{h_t}$, i.e., $D_{KL}(\mathcal T_{h_t}\|\mathcal U_{D_{h_t}})=0$, which is an optimistic situation. Then by Lemma \ref{lem-saci},
\begin{align}\label{prh-saci}
\widebar{\pr}_h \geq \frac{1-\eta}{\sum_{t=1}^{T} m_t} \sum\nolimits_{t=1}^{T} m_t \Big(\frac{|D_{\alpha^*}|(1-\epsilon_{\mathcal U_X,t})}{|D_{\alpha_t}|+\epsilon_{\mathcal U_X,t}}\Big).
\end{align}
By plugging $ \epsilon_{\mathcal U_X,t} \leq \frac{\epsilon_{\mathcal D_t}}{1-\sqrt{\frac{1}{2}D_{KL}(\mathcal D_t\| \mathcal U_X)}}=Q \cdot \epsilon_{\mathcal D_t}$, where $\epsilon_{\mathcal D_t}$ is the expected error rate of $h_t$ under the distribution $\mathcal D_t = \lambda \mathcal U_{D_{h_t}} + (1-\lambda) \mathcal U_{ X}$ and $Q=(1-\sqrt{\frac{1}{2}D_{KL}(\mathcal D_t\| \mathcal U_X)})^{-1}$,
\begin{align}\label{prh-saci-q}
\widebar{\pr}_h \geq \frac{1-\eta}{\sum_{t=1}^{T} m_t} \sum\nolimits_{t=1}^{T} m_t \Big(\frac{|D_{\alpha^*}|(1-Q\cdot\epsilon_{\mathcal D_t})}{|D_{\alpha_t}|+Q\cdot\epsilon_{\mathcal D_t}}\Big).
\end{align}

Note from Lemma \ref{lemma-classification} that, the convergence rate of the error is $\tilde O(\frac{1}{m})$ ignoring other variables and logarithmic terms from Lemma \ref{lemma-classification}. We assume that \saci{} uses learning algorithms with convergence rate $\tilde{\Theta}(\frac{1}{m})$. We then find that such \saci{} algorithms cannot exponentially improve the uniform search in the worst case, as Proposition \ref{prop-saci-impossible}.

\begin{proposition}\label{prop-saci-impossible}
Using learning algorithms with convergence rate $\tilde \Theta(\frac{1}{m})$, $\forall f,\ \alpha^*>0$ and $0<\delta<1$, with probability at least $1-\delta$, if the query complexity of the uniform search is $\texttt{superpoly}(\frac{1}{\alpha^*},n,\frac{1}{\delta})$, the query complexity of \saci{} algorithms is also $\texttt{superpoly}(\frac{1}{\alpha^*},n,\frac{1}{\delta})$ in the worst case.
\end{proposition}
\begin{myproof}
The query complexity of the uniform search being $\texttt{superpoly}(\frac{1}{\alpha^*},n,\frac{1}{\delta})$ implies that
\begin{align}
\frac{1}{\pr_u}=\frac{1}{|D_{\alpha^*}|} = \texttt{superpoly}(\frac{1}{\alpha^*},n,\frac{1}{\delta}).
\end{align}

For the \saci{} algorithms, if we ask the learning algorithm to produce a classifier with error rate $\frac{1}{\texttt{superpoly}(\frac{1}{\alpha^*},n,\frac{1}{\delta})}$, it will require  $\texttt{superpoly}(\frac{1}{\alpha^*},n,\frac{1}{\delta})$ number of samples in the worst case, so that the proposition holds. To avoid this, we can only expect the error rate to be  $\frac{1}{\texttt{poly}(\frac{1}{\alpha^*},n,\frac{1}{\delta})}$ in order to keep the query complexity at each iteration small.

Meanwhile, we can only have $T=\texttt{poly}(\frac{1}{\alpha^*},n,\frac{1}{\delta})$ iterations otherwise we will have super-polynomial number of samples.

Following the optimistic case of Eq.\eqref{prh-saci-q}, since $Q\geq 1$, we consider one more optimistic situation that $Q=1$. Let $\eta=0.5$. Even though, in the worst case that $|D_{h_t}|=|D_{\alpha_t}| + Q\epsilon_{\mathcal D_t}$,  we can have that
\begin{align}
& \widebar{\pr}_h = \frac{1}{2\sum_{t=1}^{T} m_t} \sum\nolimits_{t=1}^{T} m_t\Big(\frac{|D_{\alpha^*}|(1-\epsilon_{\mathcal D_t})}{|D_{\alpha_t}|+\epsilon_{\mathcal D_t}}\Big)\\
& = \frac{1}{\texttt{poly}(\frac{1}{\alpha^*},n,\frac{1}{\delta})} \texttt{poly}(\frac{1}{\alpha^*},n,\frac{1}{\delta}) \frac{\frac{1}{\texttt{superpoly}(\frac{1}{\alpha^*},n,\frac{1}{\delta})}}{\frac{1}{\texttt{poly}(\frac{1}{\alpha^*},n,\frac{1}{\delta})}} \\
& = \frac{1}{\texttt{superpoly}(\frac{1}{\alpha^*},n,\frac{1}{\delta})},
\end{align}
where it is noted that as long as $\epsilon_{\mathcal D_t}=\texttt{poly}(\frac{1}{\alpha^*},n,\frac{1}{\delta})$ the value of $|D_{\alpha_t}|$ cannot affect the result.
Then substituting $\widebar{\pr}_h$ into Theorem \ref{lem-sal} obtains the total samples $m_\Sigma =  \texttt{superpoly}(\frac{1}{\alpha^*},n,\frac{1}{\delta})$ that proves the proposition.
\end{myproof}

The proposition implies that the \saci{} algorithms can face the same barrier as that of the uniform search. Nevertheless, the \saci{} algorithms can still improve the uniform search within a polynomial factor. We show this by case studies.

\noindent{\bf On Sphere Function Class:} \\
Given the solution space $ X_n=\{(x_1,\ldots,x_n) \mid \forall i=1,\ldots,n: x_i \in [0, 1]\}$, the Sphere Function class is $\mathcal F^{\,n}_{sphere} = \{f^{x^*,n}_{sphere}|\forall x^*\in  X_n\}$ where
\begin{align}
f^{x^*,n}_{sphere}(x)=\frac{1}{n}\left\|x-x^{\ast}\right\|^{2}_{2}=\frac{1}{n}\sum_{i=1}^{n}{(x_i-x_i^{\ast})^2}.
\end{align}
Obviously, $|X_n|=1$, $f^{x^*,n}_{sphere}\in [0,1]$ is convex, and the optimal value is 0. It is important to notice that the volume of a $n$-dimensional hyper-sphere with radius $r$ is $\frac{\pi^{\frac{n}{2}}}{\Gamma(\frac{n}{2}+1)}r^n$, where $\Gamma(s)=\int_{0}^{\infty}{t^{s-1}}e^{-t}\,\mathrm{d}t$, so that $|D_{\alpha}|=\frac{\pi^{\frac{n}{2}}}{\Gamma(\frac{n}{2}+1)}(n\alpha)^{n/2} = C_n (\alpha)^{n/2}$ for any $\alpha>0$, where $C_n=\Theta\left({(2\pi e)^{\frac{n}{2}}}/{\sqrt{\pi n}}\right)$, since the radius leading to $f^{x^*,n}_{sphere}(x)=\frac{1}{n}\left\|x-x^{\ast}\right\|^{2}_{2}\leq \alpha$ is $\sqrt{n\alpha}$.

Note that $\pr_u = |D_{\alpha^*}| = C_n(\alpha^*)^{n/2}>(\alpha^*)^{n/2}$. It is straightforward to obtain that, minimizing any function in $\mathcal F^{\,n}_{sphere}$ using the uniform search, the \paa{} query complexity with approximation level $\alpha^*>0$ is, with probability at least $1-\delta$,
\begin{align}
 O\left( (\frac{1}{\alpha^*})^{\frac{n}{2}} \ln\frac{1}{\delta} \right).
\end{align}

We assume $\mathcal L_{sphere}$ is a learning algorithm that searches in the hypothesis space $\mathcal{H}_n$ consisting of all the hyper-spheres in $\mathbb{R}^{n}$ to find a sphere that is consistent with the training data, and meanwhile the sphere satisfies the error-target independence condition. Then a SAC algorithm using $\mathcal L_{sphere}$ is a \saci{} algorithm. We simply assume that the search of the consistent sphere is feasible. Note that $VC(\mathcal{H}_n)=n+1$.
\begin{lemma}\label{lem-transform}
For any $h_t$, denote $\epsilon_{\mathcal U_X}$ as the error rate of $h_t$ under the uniform distribution over $X$ and $\epsilon_{\mathcal D_t}$ as the error rate of $h_t$ under the distribution $\mathcal D_t=\lambda \mathcal U_{D_{h_t}} + (1-\lambda) \mathcal U_X$, then it holds that
\begin{align}
\epsilon_{\mathcal U_X} \leq \frac{1}{1-\lambda} \epsilon_{\mathcal D_t},
\end{align}
where $\lambda\in[0,1]$ and $\mathcal U_{D_{h_t}}$ is the uniform distribution over $D_{h_t}$.
\end{lemma}
\begin{myproof}
Let $I[\cdot]$ be the indicator function and $D_{\neq}$ be the area where $h_t$ makes mistakes. We split $D_{\neq}$ into $D_{\neq}^+ = D_{\neq}\cap D_{h_t}$ and $D_{\neq}^-= D_{\neq} \setminus D_{\neq}^+$. We can calculate the probability density that $\mathcal D_t(x) = \lambda \frac{1}{|D_{h_t}|}+(1-\lambda)\frac{|D_{h_t}|}{| X|}\frac{1}{|D_{h_t}|}$ for any $x\in D_{\neq}^+$, and $\mathcal D_t(x) = (1-\lambda)\frac{| X\setminus D_{h_t}|}{| X|}\frac{1}{| X\setminus D_{h_t}|} = (1-\lambda)\frac{1}{| X|}$ for any $x\in D_{\neq}^-$. Thus,
\begin{align}
	\epsilon_{\mathcal D_t} & = \int_{ X} \mathcal D_t(x) I[h_t\text{ makes mistake on } x] \mathrm{d}x\\
	& = \int_{D_{\neq}} \mathcal D_t(x)  \mathrm{d}x = \int_{D_{\neq}^+} \mathcal D_t(x)  \mathrm{d}x + \int_{D_{\neq}^-} \mathcal D_t(x)  \mathrm{d}x\\
	& \geq \int_{D_{\neq}^+} (1-\lambda)\frac{1}{| X|}  \mathrm{d}x + \int_{D_{\neq}^-} (1-\lambda)\frac{1}{| X|}  \mathrm{d}x\\
	& = (1-\lambda)\epsilon_{\mathcal U_X},
\end{align}
which proves the lemma.
\end{myproof}
We then obtain the \paa{} complexity as in Proposition \ref{prop-saci-sphere}.

\begin{proposition}\label{prop-saci-sphere}
For any function in $\mathcal F^{\,n}_{sphere}$ and any approximation level $\alpha^*>0$, \saci{}  algorithms can achieve the \paa{} query complexity, for any $n\geq 2$,
\begin{align}
 O\left((\frac{1}{\alpha^*})^{\frac{n-1}{2}} \log\frac{1}{\sqrt{\alpha^*} } (\ln\frac{1}{\delta}+n\log\frac{1}{\sqrt{\alpha^*} })\right) 
\end{align}
with probability at least $1-\delta$.
\end{proposition}
\begin{myproof}
We choose $\alpha_t=\frac{1}{2^t}$ for all $t$, and use the number of iterations $T$ to approach $|D_{\alpha_T}|=\sqrt{|D_{\alpha^*}|}$, for the approximation level $\alpha^*$. Solving this equation with the sphere volume results in $T=\log\frac{(C_n)^{\frac{1}{n}}}{\sqrt{\alpha^*}}$. We let the \saci{} algorithm run $T=\log\frac{1}{\sqrt{\alpha^*}}$ number of iterations. We assume $\log\frac{1}{\sqrt{\alpha^*}}$ is an integer for simplicity, which does not affect the generality.

In iteration $t$, using $\mathcal L_{sphere}$, we want the error of the hypothesis $h_t$, $\epsilon_{\mathcal D_t}$, to be $\frac{1}{2^t}$. Since the $\mathcal L_{sphere}$ produces a hypothesis with zero training error, from
\begin{align}
\epsilon_{\mathcal D_t}=\frac{1}{2^t} \leq 2m^{-1}\left(d\log{(2emd^{-1})}+\log{(2\eta^{-1})}\right),
\end{align}
we can solve the required sample size with $\eta$ being a constant,
\begin{align}
m_t \leq m_T = O(n T 2^{T}) = O\left(\frac{n}{\sqrt{\alpha^*}}\log\frac{1}{\sqrt{\alpha^*}}\right)
\end{align}
using the inequality $\log{x} \leq cx-(\log{c}+1)$ for any $x>0$ and any $c>0$. We thus obtain $\sum_{t=1}^{T}m_t=O\left(\frac{n}{\sqrt{\alpha^*}}(\log\frac{1}{\sqrt{\alpha^*}})^2\right)$.

We then follow Eq.\eqref{prh-saci-q}. We use uniform sampling within $D_{h_t}$, then $Q=\frac{1}{1-\lambda}$. Letting the \saci{} algorithms use $m_T$ number of samples in every iteration, $\lambda=0.5$ and $\eta=0.5$, we have
\begin{align}
	&\widebar{\pr}_h \geq  \frac{1}{ 2 \log \frac{1}{\sqrt{\alpha^*}} } \sum\nolimits_{t=1}^{\log \frac{1}{\sqrt{\alpha^*}}} \Big(\frac{|D_{\alpha^*}|(1-Q\epsilon_{\mathcal D_t})}{|D_{\alpha_t}|+Q\epsilon_{\mathcal D_t}}\Big)\\
	& \geq  \frac{C_n(\alpha^*)^\frac{n}{2}}{ 2 \log \frac{1}{\sqrt{\alpha^*}} } \sum\nolimits_{t=1}^{\log \frac{1}{\sqrt{\alpha^*}}}{\frac{1-2\frac{1}{2^t}} {C_n(\frac{1}{2^t})^\frac{n}{2}+2\frac{1}{2^t}}}\\
	& \geq  \frac{C_n(\alpha^*)^\frac{n}{2}}{ 2 \log \frac{1}{\sqrt{\alpha^*}} } \frac{1}{2(C_n+2)}\sum\nolimits_{t=2}^{\log{\frac{1}{\sqrt{\alpha^*}}}}{2^t}\\
	& =  \frac{C_n(\alpha^*)^\frac{n}{2}}{ 2 \log \frac{1}{\sqrt{\alpha^*}} }\frac{(\frac{1}{\sqrt{\alpha^*}}-2 )}{(C_n+2)}= \Omega\Big(\frac{(\alpha^*)^\frac{n-1}{2}}{ \log \frac{1}{\sqrt{\alpha^*}} }\Big).
\end{align}
So we obtain the query complexity from Theorem \ref{lem-sal}
\begin{align}
O\left(m_0+\max\Big\{(\frac{1}{\alpha^*})^{\frac{n-1}{2}} \log\frac{1}{\sqrt{\alpha^*} } \ln\frac{1}{\delta},  \frac{n}{\sqrt{\alpha^*}}(\log\frac{1}{\sqrt{\alpha^*}})^2\Big\}\right)
\end{align}
which is $O\left((\frac{1}{\alpha^*})^{\frac{n-1}{2}} \log\frac{1}{\sqrt{\alpha^*} } (\ln\frac{1}{\delta}+n\log\frac{1}{\sqrt{\alpha^*} })\right)$ using a constant $m_0$ and the $\max$ is upper bounded by plus.
\end{myproof}

We can see that the \saci{} algorithms can accelerate the uniform search by a factor near $\frac{1}{\sqrt{\alpha^*}}/\log \frac{1}{\sqrt{\alpha^*}}$. The closer the approximation, the more the acceleration.

\noindent{\bf On Spike Function Class}\\
As modeling EAs, SAL algorithms should be expected to be applied on the complex problems, while the Sphere Function class only consists of convex functions. Inherited from EAs, SAL algorithms can handle problems with some local optima. We show this by comparing \saci{} with the uniform search on the Spike Function class defined below.

Define regions $A_{1,k}= [\frac{3k}{20}, \frac{3k+2}{20}]$ where $0 \leq k\in \mathbb{N} \leq 6$ and $A_{2,k}= (\frac{3k-1}{20}, \frac{3k}{20})$ where $1 \leq k\in \mathbb{N} \leq 6$, and define $g(x)$ over $[0,1]$ that
\begin{align}
g(x)=
\begin{cases}
	x-\frac{k}{10}, & x\in A_{1,k}\\
	-x+\frac{k}{5}, & x\in A_{2,k}
\end{cases}
\end{align}

Let $ X_n = [-\frac{1}{2},\frac{1}{2}]^n$ be the $n$-dimensional solution space. The Spike Function class is $\mathcal F^n_{spike}=\{f^{x^*,n}_{spike}|\forall x^*\in  X_n \}$, where, for all $x \in X_n$
\begin{align}
f^{x^*,n}_{spike}(x) = g(\frac{1}{\sqrt{n}}\|x-x^*\|_2).
\end{align}
It is easy to know $\min_{x \in X_n} f(x)=0$ and $\max_{x \in X_n} f(x)\leq 1$ for any $f\in\mathcal F^n_{spike}$. For any $\alpha>0$, we can bound the area $|D_\alpha| \in [C_n \alpha^{n}, C_n (3\alpha)^{n}]$, where $C_n=\Theta\left({(2\pi e)^{\frac{n}{2}}}/{\sqrt{\pi n}}\right)$.

\begin{figure}[t]
\centering
\includegraphics[width=0.85\linewidth]{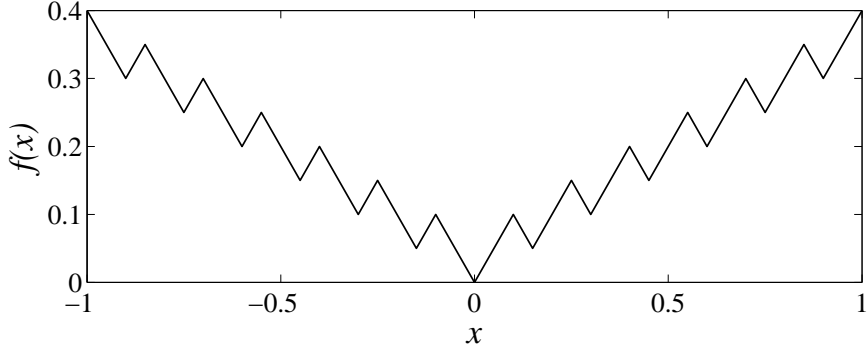}
\caption{The landscape of function $f^{0,1}_{spike}(x)$ in $[-1,1]$.
}
\label{fig-example}
\end{figure}

The Spike functions are non-convex and non-differentiable with some local optima, as depicted in Figure \ref{fig-example}.

Minimizing any function in $\mathcal F^n_{spike}$ using the uniform search, the \paa{} query complexity with approximation level $\alpha^*>0$ is, with probability at least $1-\delta$,
\begin{align}
 O\left( (\frac{1}{\alpha^*})^{n} \ln\frac{1}{\delta} \right).
\end{align}

We configure the \saci{} algorithm to use the learning algorithm $\mathcal L_{spike}$ that searches the smallest sphere covering all the samples labeled as positive, of which the VC-dimension is $n+1$. Note that since the function is non-convex, the $\mathcal L_{spike}$ may output a sphere that also covers some negative examples, and thus with some training error. Using this \saci{} algorithm to minimize any member in the function class $\mathcal{F}^n_{spike}$, we obtain the \paa{} query complexity as in Proposition \ref{prop-spike}.
\begin{proposition}\label{prop-spike}
For any function in $\mathcal{F}^n_{spike}$ and any approximation level $\alpha^*>0$, \saci{} algorithms can achieve the \paa{} query complexity
\begin{align}
 O\left((\frac{1}{\alpha^*})^{n-\frac{1}{2}} \log\frac{1}{\sqrt{\alpha^*} } \big(\ln\frac{1}{\delta} + n\log\frac{1}{\sqrt{\alpha^*}}\big)\right),
\end{align}
with probability at least $1-\delta$.
\end{proposition}
\begin{myproof}
For any function in $\mathcal{F}^n_{spike}$, we note that the function is convex in $D_\alpha$ when $\alpha$ is smaller than $0.05$. We set $\alpha_t=\frac{1}{2^t}$, so that when $t\geq 5$, the \saci{} algorithm with $\mathcal L_{spike}$ will deal with a convex function and thus the training error is zero. We use the number of iterations $T$ to achieve $|D_{\alpha_T}|=\sqrt{|D_{\alpha^*}|}$. Since $|D_\alpha| \in [C_n \alpha^{n}, C_n (3\alpha)^{n}]$, we can obtain $T\geq \log\frac{3(C_n)^\frac{1}{2n}}{\sqrt{\alpha^*}}$. We let the \saci{} algorithm run $T=\log\frac{1}{\sqrt{\alpha^*}}$ number of iterations and assume that $\log\frac{1}{\sqrt{\alpha^*}}$ is an integer.

In iteration $t\geq 5$, we want the error of the hypothesis $h_t$, $\epsilon_{\mathcal D_t}$, to be $\frac{1}{2^t}$. Since the training error can be zero, we can solve the required sample size
$
m_t\leq m_T = O\left(\frac{n}{\sqrt{\alpha^*}}\log \frac{1}{\sqrt{\alpha^*}}\right)
$. We thus obtain $\sum_{t=1}^{T}m_t=O\left(\frac{n}{\sqrt{\alpha^*}}(\log\frac{1}{\sqrt{\alpha^*}})^2\right)$.

We then follow Eq.\eqref{prh-saci-q}. We use uniform sampling within $D_{h_t}$, then $Q=\frac{1}{1-\lambda}$. Letting the \saci{} algorithm use $m_T$ number of samples in every iteration, $\lambda=0.5$ and $\eta=0.5$, we have
\begin{align}
	&\widebar{\pr}_h \geq \frac{1}{ 2\log \frac{1}{\sqrt{\alpha^*}} } \sum\nolimits_{t=5}^{\log \frac{1}{\sqrt{\alpha^*}}} \Big(\frac{|D_{\alpha^*}|(1-Q\epsilon_{\mathcal D_t})}{|D_{\alpha_t}|+Q\epsilon_{\mathcal D_t}}\Big)\\
	& \geq  \frac{C_n(\alpha^*)^n}{ 2\log \frac{1}{\sqrt{\alpha^*}} } \sum\nolimits_{t=5}^{\log \frac{1}{\sqrt{\alpha^*}}}{\frac{1-2\frac{1}{2^t}} {C_n(\frac{3}{2^t})^n +2\frac{1}{2^t}}}\\
	& \geq  \frac{C_n(\alpha^*)^n}{ 2\log \frac{1}{\sqrt{\alpha^*}} } \frac{ \frac{15}{16} }{3C_n+2} \sum\nolimits_{t=5}^{\log \frac{1}{\sqrt{\alpha^*}}} 2^t\\
	& = \frac{15}{32} \frac{ C_n(\alpha^*)^n}{ \log \frac{1}{\sqrt{\alpha^*}} } \frac{\frac{2}{\sqrt{\alpha^*}}-2^5}{3C_n+2} = \Omega\Big(\frac{(\alpha^*)^{n-\frac{1}{2}}}{ \log \frac{1}{\sqrt{\alpha^*}} }\Big).
\end{align}
So we obtain the query complexity from Theorem \ref{lem-sal}
\begin{align}
O\left((\frac{1}{\alpha^*})^{n-\frac{1}{2}} \log\frac{1}{\sqrt{\alpha^*} } \big(\ln\frac{1}{\delta} + n\log\frac{1}{\sqrt{\alpha^*}}\big)\right)
\end{align}
as the $\max$ is upper bounded by plus.
\end{myproof}

We observe from the proof that the non-convexity can result in non-zero training error for the learning algorithms in \saci{} algorithms, and thus the search process is interfered. But as long as the non-convexity is not quite severe, like the Spike Functions, \saci{} algorithms are not significantly affected, and can still be better than the uniform search by a factor near $\frac{1}{\sqrt{\alpha^*}}/\log \frac{1}{\sqrt{\alpha^*}}$.

\subsection{A Super-Polynomial Acceleration Condition}

We have shown in Proposition \ref{prop-saci-impossible} that \saci{} algorithms using common classification algorithms cannot super-polynomially improve from the uniform search in the worst case. An interesting question is therefore raised that when the super-polynomial improvement is possible.

Learned from the proof of Proposition \ref{prop-saci-impossible}, a straightforward way is to use a powerful classification algorithm with exponentially improved sample complexity, i.e., $\tilde O(\ln \frac{1}{\epsilon})$, so that only a polynomial number of samples is required to achieve a super-polynomially small error. Several \emph{active learning} algorithms can do this in some circumstances (e.g. \cite{DasguptaKM09,WangZ10}). Applying active learning algorithms needs a small modification of \saci{}. In iteration $t$, instead of sampling from the uniform distribution in $D_{h_t}$, the sampling is guided by the classifier. Nevertheless, the achieved error is still evaluated under the original (uniform) distribution. Using such learning algorithms denoted as $\mathcal L^{\ln}_{sphere}$, we achieve Proposition \ref{prop-saci-sphere-ln} showing a super-polynomial acceleration from the uniform search on Sphere Functions.

\begin{proposition}\label{prop-saci-sphere-ln}
For any function in $\mathcal F^{\,n}_{sphere}$ and any approximation level $\alpha^*>0$, \saci{}  algorithms using $\mathcal L^{\ln}_{sphere}$ can achieve the \paa{} query complexity, for any $n\geq 2$,
\begin{align}
O\left(\log\frac{1}{\alpha^* }(\ln\frac{1}{\delta}+n \log\frac{1}{\alpha^* })\right)
\end{align}
with probability at least $1-\delta$.
\end{proposition}
\begin{myproof}
We choose $\alpha_t=\frac{1}{2^t}$ for all $t$, and use the number of iterations $T$ to approach $|D_{\alpha_{T}}|=|D_{\alpha^*}|$, for the approximation level $\alpha^*$. Solving this equation with the sphere volume results in $T=\log\frac{(C_n)^{\frac{1}{n}}}{\alpha^*}$. We let the \saci{} algorithm run $T=\log\frac{1}{\alpha^*}$ number of iterations. We assume $\log\frac{1}{\alpha^*}$ is an integer for simplicity, which does not affect the generality.

In iteration $t$, using $\mathcal L^{\ln}_{sphere}$, we want the error of the hypothesis $h_t$, $\epsilon_{\mathcal D_t}$, to be $\frac{1}{2^{tn/2}}$. Since the $\mathcal L^{\ln}_{sphere}$ has the sample complexity $O(\ln \frac{1}{\epsilon})$, we ask for a hypothesis with zero training error, which requires the sample size $m_t=O(tn) = O(n\log\frac{1}{\alpha^*})$ with $\eta$ being a constant.

We thus obtain $\sum_{t=1}^{T}m_t=O\left(n\big(\log\frac{1}{\alpha^*}\big)^2\right)$.

Following Eq.\eqref{prh-saci-q}, we use uniform sampling within $D_{h_t}$, then $Q=\frac{1}{1-\lambda}$. Letting the \saci{} algorithms use $m_T$ number of samples in every iteration, $\lambda=0.5$ and $\eta=0.5$, we have
\begin{align}
	\widebar{\pr}_h & \geq  \frac{1}{ 2\log \frac{1}{\alpha^*} } \sum\nolimits_{t=1}^{\log \frac{1}{\alpha^*}} \Big(\frac{|D_{\alpha^*}|(1-Q\epsilon_{\mathcal D_t})}{|D_{\alpha_t}|+Q\epsilon_{\mathcal D_t}}\Big)\\
	& \geq  \frac{C_n(\alpha^*)^\frac{n}{2}}{ 2\log \frac{1}{{\alpha^*}} } \sum\nolimits_{t=1}^{\log \frac{1}{\alpha^*}}{\frac{1-2(\frac{1}{2^t})^{\frac{n}{2}}} {C_n(\frac{1}{2^t})^\frac{n}{2}+2(\frac{1}{2^t})^{\frac{n}{2}}}}\\
	& \geq  \frac{C_n(\alpha^*)^\frac{n}{2}}{ 2\log \frac{1}{{\alpha^*}} } \frac{1}{2(C_n+2)}\sum\nolimits_{t=2}^{\log{\frac{ 1 }{\alpha^*}}}{\frac{1} {(\frac{1}{2^t})^{\frac{n}{2}}}}\\
	& \geq  \frac{C_n(\alpha^*)^\frac{n}{2}}{ 2\log \frac{1}{{\alpha^*}} }\frac{(\frac{1}{\alpha^*})^{\frac{n}{2}} \left(1-(2\alpha^*)^\frac{n}{2}\right) } {2(C_n+2)}
	= \Omega\Big(\frac{1}{ \log \frac{1}{{\alpha^*}} }\Big).
\end{align}
So we obtain the query complexity from Theorem \ref{lem-sal}, letting $m_0$ be a constant,
\begin{align}
O\left(\max\Big\{ \log\frac{1}{\alpha^* } \ln\frac{1}{\delta},  n(\log\frac{1}{\alpha^*})^2\Big\}\right)
\end{align}
which is $O\left( \log\frac{1}{\alpha^* }(\ln\frac{1}{\delta} + n \log\frac{1}{\alpha^* }) \right)$.
\end{myproof}

Meanwhile, we are more interested in exploring conditions under which the super-polynomial improvement is possible without requiring such powerful learning algorithms. For this purpose, we find the \emph{one-side-error} condition.

\begin{cond}[One-Side-Error] In SAC algorithms, for any $t$ and any $x\in X$, if $x\in D_{h_t}\Delta D_{\alpha_t}$, it must hold that $x\in D_{\alpha_t}$.
\end{cond}

The condition implies that $h_t$ can only make false-negative errors, i.e., wrongly classifies positive samples (inside $D_{\alpha_t}$) as negative, but no false-positive errors. One practical way to approach this condition is through the cost-sensitive classifiers \cite{DBLP:conf/ijcai/Elkan01,zhou.liu.tkde06} with a very large mis-classification cost for negative samples.  We call \saci{} algorithms that are further under this condition as \sacii{} algorithms.

\begin{lemma}\label{lem-sacii}
For \sacii{} algorithms, it holds for all $t$ that
$
 |D_{h_t}|\leq |D_{\alpha_{t}}|.
$
\end{lemma}
\begin{myproof}
Note that for training $h_t$ we label the samples from $D_{\alpha_t}$ as positive and label the rest as negative. Since $h_t$ only makes false-negative errors, i.e., every error is in $D_{\alpha_t}$, we have $D_{h_t}\subseteq D_{\alpha_t}$, which implies the lemma.
\end{myproof}

Lemma \ref{lem-sacii} shows that the one-side-error condition controls the size $|D_{h_t}|$ to be bounded by $|D_{\alpha_t}|$. Thus we can refine Lemma \ref{lem-saci} as Lemma \ref{lem-sacii-2}.

\begin{lemma}\label{lem-sacii-2}
For \sacii{} algorithms, it holds for all $t$ that
\begin{align}
\frac{|D_{\alpha^*}\cap D_{h_t}|}{|D_{h_t}|} \geq \frac{|D_{\alpha^*}|(1-\epsilon_{\mathcal U_X,t})}{|D_{\alpha_t}|},
\end{align}
where $\epsilon_{\mathcal U_X,t}$ is the expected error rate of $h_t$ under $\mathcal U_X$.
\end{lemma}
\begin{myproof}
Since the \sacii{} algorithm is also a \saci{} algorithm, incorporating Lemma \ref{lem-sacii} into Lemma \ref{lem-saci} proves the lemma.
\end{myproof}

We assume that $\mathcal L^{+}_{sphere}$ is a learning algorithm that not only behaviors like $\mathcal L_{sphere}$ but also results a hypothesis satisfying the one-side-error condition. Then a \saci{} algorithm using $\mathcal L^{+}_{sphere}$ is a \sacii{} algorithm. We again assume that $\mathcal L^{+}_{sphere}$ is feasible, of which $VC(\mathcal{H}_n)=n+1$. We then use this \sacii{} algorithm on the Sphere Function class, on which \saci{} algorithms bear a super-polynomial \paa{} complexity, and obtain Proposition \ref{prop-sacii-sphere}.


\begin{proposition}\label{prop-sacii-sphere}
For any function in $\mathcal F^{\,n}_{sphere}$ and any approximation level $\alpha^*>0$, \sacii{}  algorithms can achieve the \paa{} query complexity
\begin{align} O\left(\log\frac{1}{\alpha^* } (\ln\frac{1}{\delta}+n)\right), 
\end{align}
with probability at least $1-\delta$.
\end{proposition}
\begin{myproof}
By Lemma \ref{lem-sacii-2},
\begin{align}
	\frac{|D_{\alpha^*} \cap D_{h_t}|}{|D_{h_{t}}|} 
	& \geq \frac{|D_{\alpha^*}|(1 - Q\epsilon_{\mathcal D_t} )}{|D_{\alpha_t}|},
\end{align}
where $\epsilon_{\mathcal D_t}$ is the error of $h_t$ under its original distribution $\mathcal D_t$, and $Q$ is the resulting factor of changing the distribution.

Let $\alpha_t=\frac{1}{2^t}$ for all $t$, and use the number of iterations $T$ to achieve $|D_{\alpha_{T}}|=|D_{\alpha^*}|$, for the approximation level $\alpha^*$. Solving this equation with the sphere volume results in $T=\log\frac{(C_n)^{\frac{1}{n}}}{\alpha^*}$. We let the \sacii{} algorithm run $T=\log\frac{1}{\alpha^*}$ number of iterations. We assume $\log\frac{1}{\alpha^*}$ is an integer for simplicity, which does not affect the generality.

In iteration $t$, using $\mathcal L^+_{sphere}$, we want the error of the hypothesis $h_t$, $\epsilon_{\mathcal D_t}$, to be a constant $\frac{1}{2}$. Since $\mathcal L^+_{sphere}$ produces a hypothesis with zero training error, to achieve $\epsilon_{\mathcal D_t}\leq \frac{1}{2}$ it requires the number of samples in $O(n)$.
We thus obtain $\sum_{t=1}^{T}m_t=O\left(n\log\frac{1}{\alpha^*}\right)$.

We then follow Eq.\eqref{prh-saci-q}. We use uniform sampling within $D_{h_t}$, then $Q=\frac{1}{1-\lambda}$. Letting the \sacii{} algorithm use $m_T$ number of samples in every iteration, $\lambda=\frac{1}{3}$ and $\eta=0.5$, we have
\begin{align}
	&\widebar{\pr}_h \geq  \frac{1}{ 2\log \frac{1}{\alpha^*} } \!\sum\nolimits_{t=1}^{\log \frac{1}{\alpha^*}} \!\!\Big(\frac{|D_{\alpha^*}|(1 - Q \epsilon_{\mathcal D_t} )}{|D_{\alpha_t}|}\Big)\!\\
	& \geq  \frac{1}{ 2\log \frac{1}{\alpha^*} } \sum\nolimits_{t=1}^{\log \frac{1}{\alpha^*}} \Big(\frac{\frac{1}{4}|D_{\alpha^*}|}{|D_{\alpha_t}|}\Big)\\
	& =  \frac{C_n(\alpha^*)^\frac{n}{2}}{ 8 \log \frac{1}{\alpha^*} } \sum\nolimits_{t=1}^{\log \frac{1}{\alpha^*}} \frac{1}{C_n (\frac{1}{2^{t}})^{\frac{n}{2}}}\\	
	& \geq  \frac{C_n(\alpha^*)^\frac{n}{2}}{ 8 \log \frac{1}{\alpha^*} }  \frac{ \left( (\frac{1}{\alpha^*})^{\frac{n}{2}}-1 \right) }{C_n}
	= \Omega\Big(\frac{1}{\log \frac{1}{\alpha^*} }\Big).
\end{align}
So we obtain from Theorem \ref{lem-sal} the query complexity of the \sacii{} algorithm, letting $m_0$ be a constant,
\begin{align}
O\left(\max\left\{\log\frac{1}{\alpha^* }\ln\frac{1}{\delta},  n\log\frac{1}{\alpha^* }\right\}\right),
\end{align}
which is $O\left(\log\frac{1}{\alpha^* } (\ln\frac{1}{\delta}+n)\right)$.
\end{myproof}

Proposition \ref{prop-sacii-sphere} shows a super-polynomial improvement from the complexity of the uniform search. It is interesting to note that we only ask for a random guess classification (i.e., error rate $\frac{1}{2}$) in the proof of Proposition \ref{prop-sacii-sphere}.

\section{Discussions and Conclusions}

This paper describes the sampling-and-learning (SAL) framework which is an abstract summary of a range of EAs. The SAL framework allows us to investigate the general performance of EAs from a statistical view. We show that the SAL framework directly admits a general upper bound on the \paa{} query complexity, which is the number of fitness evaluations before an approximate solution is found with a probability.

Focusing on SAC algorithms, which are SAL algorithms using classification learning algorithms, we give a more specific performance upper bound, and compare with uniform random search. We find two conditions that drastically effect the performance of SAC algorithms. Under the \emph{error-target independence} condition, which assumes that the error of the learned classifier in each iteration is independent with the target approximation area, the SAC algorithms can obtain a polynomial improvement over the uniform search, but not a super-polynomial improvement. We demonstrate the improvement using the Sphere Function class consisting of convex functions as well as the Spike Function class consisting of non-convex functions. Further incorporating the \emph{one-side-error} condition, which assumes that the classification only makes false-negative errors, the SAC algorithms can obtain a super-polynomial improvement over the uniform search.

On the one hand, our results show that the property of classification error in SAC algorithms greatly impacts the performance, which was never touched in previous studies, as far as we know. We expect the work could guide the design of novel search algorithms. On the other hand, how to satisfy the conditions is a non-trivial practical issue.

In the case study on the Sphere Function class, we find that a learning error rate no more than the random guess is sufficient to achieve a super-polynomial improvement under the conditions. This implies that an accurate learning algorithm may not be necessary for a good SAC algorithm. It is interesting that a recent work \cite{Bei.etal.stoc13} also noticed that a learnable concept is not necessary for the trial-and-error search with a computation oracle.

In this paper, the SAC algorithms are analyzed in continuous domains, while the main body of theoretical studies of evolutionary algorithms focuses on the discrete domains. Thus understanding the performance of SAC algorithms in discrete domains is our future work. Moreover, in the SAC algorithms analyzed in this paper, the learning algorithm does not utilize the last hypothesis or the last data set. It would be interesting to investigate whether considering them will bring any significant difference.

\section*{Acknowledgements}

This work was supported by the National Science Foundation of China (61375061) and the Jiangsu Science Foundation (BK2012303).

\bibliography{ectheory}
\bibliographystyle{abbrvnat}
\end{document}